\documentclass{article}
\usepackage{Template/ijcai18}

\usepackage{times}
\usepackage{subcaption}
\usepackage{url}
\usepackage{amsthm}
\usepackage{scalerel}

\usepackage{booktabs}
\usepackage{tabularx}
\usepackage[font=small]{caption}
\usepackage{listings}
\usepackage{xcolor}
\usepackage{xspace}
\usepackage{paralist}
\usepackage{amsmath,amsfonts,amssymb,stmaryrd}
\usepackage{bigints}
\usepackage{graphicx}
\usepackage{amsthm}
\usepackage{oz}
\usepackage{ifthen}
\usepackage{tabularx}
\usepackage{helvet}
\usepackage{algpseudocode}
\usepackage{algorithm}
\usepackage{algorithmicx}
\usepackage{multirow}

\algrenewcommand\alglinenumber[1]{\tiny #1:}

\lstnewenvironment{limamodel}[1][]
    {\lstset{float=tbp,basicstyle=\footnotesize\ttfamily,#1}}
    {}

\newcommand\multiset[1]{\ensuremath{\llbracket\, #1 \,\rrbracket}}

\newcommand{\densitylabel}[1]{\ensuremath #1}
\newcommand{\density}[2]{\ensuremath{\densitylabel{\mbox{\ensuremath{ #1:\ }}} #2}}

\newcommand{\slotn}[1]{\mbox{#1}}
\newcommand{\slot}[2]{\ensuremath{\slotn{#1:\ }\densitylabel{#2}}}

\newcommand{\entity}[2][-1]{
  \ifthenelse{\equal{#1}{-1}}{
    \ensuremath{\langle #2 \rangle}
  }{
    \ensuremath{#1 \langle #2 \rangle}
  }
}

\newcommand{\context}[1]{\ensuremath{\langle #1 \rangle}}

\DeclareMathOperator*{\foo}{\scalerel*{\uplus}{\sum}}
\DeclareMathOperator*{\myprod}{\scalerel*{\Pi}{\sum}}

\title{Lifted Filtering via Exchangeable Decomposition}

\author{
Stefan L\"udtke$^1$, 
Max Schr\"oder$^1$, 
Sebastian Bader$^1$,
Kristian Kersting$^2$,
Thomas Kirste$^1$ 
\\ 
$^1$Institute of Computer Science, University of Rostock, Germany\\
$^2$Computer Science Department and Centre for Cognitive Science, TU Darmstadt, Germany\\
\{stefan.luedtke2, max.schroeder, sebastian.bader, thomas.kirste\}@uni-rostock.de\\
 kersting@cs.tu-darmstadt.de%, \{first.last\}@uni-rostock.de, 
}
\begin{document}

\maketitle

\begin{abstract}
We present a model for exact recursive Bayesian filtering based on lifted multiset states. Combining multisets with lifting makes it possible to simultaneously exploit multiple strategies for reducing inference complexity when compared to list-based grounded state representations. The core idea is to borrow the concept of Maximally Parallel Multiset Rewriting Systems and to enhance it by concepts from Rao-Blackwellization and Lifted Inference, giving a representation of state distributions that enables efficient inference. 
In worlds where the random variables that define the system state are exchangeable -- where the identity of entities does not matter -- it automatically uses a representation that abstracts from ordering (achieving an exponential reduction in complexity) -- and it automatically adapts when observations or system dynamics destroy exchangeability by breaking symmetry. 
\end{abstract}

% !TEX root = ../P20171110-IJCAI_LiMa-SL.tex

\newcommand\mymid{\!\mid\!}

\section{Introduction}
\label{sec:intro}
Modeling dynamical systems is fundamental for the understanding of complex phenomena in a variety of AI tasks. \emph{Multiset Rewriting Systems (MRSs)} provide a convenient mechanism to represent dynamic systems that consist of multiple interacting entities which can be grouped into ``species''. 
MRSs can be used to model biochemical reactions \cite{barbuti_maximally_2011}, population dynamics in ecological studies \cite{pescini_dynamical_2006}, network protocols \cite{Cervesato:1999:MPA:794199.795111}, etc.
Or consider \emph{human activity recognition (HAR)} \cite{bulling2014tutorial}, the target of our present paper, and assume that we are interested in reconstructing structured activities of one or more human protagonists (pursuing everyday activities) from noisy or ambiguous sensor data. 
Here, multisets naturally arise when we want to represent states that encompass multiple persons, or multiple objects that are handled by persons, which can not be discriminated by observations. 
In HAR, one established method for deriving state distributions from sequences of observations is \emph{recursive Bayesian state estimation} (RBSE), for instance based on various kinds of enhanced hidden Markov models \cite{bui2002policy,liao2007learning} or particle filtering \cite{fox2003bayesian,kruger_computational_2014}. 
RBSE iteratively computes the posterior $p(S_t \mymid y_{1:t})$ for time $t$ from the previous posterior $p(S_{t-1} \mymid y_{1:t-1} )$ at time $t - 1$ and an observation $y_t$.

In application domains such as chemistry, cell biology or ecology, MRSs are typically used for simulation studies. However, MRSs should in principle also allow the online integration of sensor data as required by HAR and as provided by recursive Bayesian state estimation. However, when we try to apply established methods to systems whose dynamics is represented by MRSs, we find that they cannot efficiently represent the exact posterior, in terms of the amount of storage required.

In this paper, we present a novel algorithm for exact RBSE for systems whose state space and dynamics can be represented by MRSs, which we call \emph{Lifted Marginal Filtering} (LiMa). The central technical idea of LiMa is to introduce a suitable factorized representation of distributions over multisets which allows to represent some factors in a parametric way rather than by samples or by complete enumeration, similar to \emph{Rao-Blackwellization} as used in particle filters \cite{doucet_rao-blackwellised_2000}. 
Interestingly, the prediction and update steps of RBSE can be performed directly on the factorized representation, without resorting to the original, much larger distribution, by exploiting
exchangeability \cite{niepert2014tractability}.
In certain cases (computing applicable actions, and when conditioning on \emph{identifying} observations), the representation of the factorized distribution needs to be manipulated by \emph{splitting} operations, similar to splitting used in exact lifted inference algorithms
 \cite{poole_first-order_2003}.

Depending on the underlying model, this approach reduces the number of explicitly represented states $s_t$ by orders of magnitude in comparison to an approach using samples or enumeration of the $s_t$.
This is possible because LiMa combines two effects: The multiset state representation, which allows to exploit exchangeability (Lifted Inference), and the factorization of the multiset state distribution, which allows manipulating the state distribution on a parametric level (Rao-Blackwellization). 
To the best of our knowledge, this is the first attempt to provide RBSE for systems with MRS dynamics and the first attempt that allows to perform prediction and update directly (at least partially) on the factorized representation.

The paper is organized as follows:
We continue by briefly introducing \emph{Probabilistic Maximally Parallel Multiset Rewriting Systems (PMPMRS)} in Section~\ref{sec:background}.
The factorization of distributions over multisets is introduced in Section~\ref{sec:factorisation}. 
In Section~\ref{sec:rbse}, we show how RBSE can be performed on the factorized representation, without needing to enumerate the original states. 
Our approach is demonstrated on two multi-agent activity recognition tasks in Section~\ref{sec:evaluation}.

% !TEX root = ../P20171110-IJCAI_LiMa-SL.tex

\section{Background}
\label{sec:background}
In the following, we briefly introduce some background on multiset rewriting systems (MRSs).

\newcommand{\myparagraph}[1]{\par\smallskip\noindent\textbf{#1}}

\newcommand{\uminus}{{\kern 2pt \cup \kern -5pt \text{-} \kern 4pt}}
\myparagraph{Multisets:}
Let $\mathcal{E}$ be a set of species. 
A multiset (over $\mathcal{E}$) is a map $s : \mathcal{E} \to \mathbb{N}$ from species to multiplicities (natural numbers). 
Let $s$, $s'$ be two multisets, let multiset union $s \uplus s'$, multiset difference $s \uminus s'$ and multiset subset relation $s \sqsubseteq s'$ be defined the obvious way. 
For species $s_1 , \dots , s_k \in \mathcal{E}$ and multiplicities $n_1 ,\dots , n_k \in \mathbb{N}$ where $n_i > 0$ we write $\multiset{n_1 e_1,\dots ,n_k e_k}$ to denote a multiset, where the multiplicity of $e_i$ is $n_i$ and  the multiplicities of all species not listed is $0$.

\myparagraph{Actions:}
For now, it is sufficient to consider a rewriting rule as a pair of multisets $(p, e)$, where $p$ are the prerequisites and $e$ is the add list. 
The action (also known as \emph{rewriting rule}) $(p,e)$ is applicable to a multiset $m$ if $p \sqsubseteq m$. 
The result of applying $(p,e)$ to m is the multiset $(m \uminus p) \uplus e$.

\myparagraph{Compound Actions:}
In scenarios where multiple entities (inter-)act simultaneously, multiple actions may take place between consecutive time steps. 
This intuition is captured by \emph{compound actions}, that describe the transition semantics of \emph{maximally parallel multiset rewriting systems} (MPMRSs) \cite{barbuti_maximally_2011}:
Each individual takes part in an action if possible, and all actions are performed in parallel.
Specifically, a compound action is a multiset of actions. 
The compound action $c$ is \emph{applicable} in a state $s$ if all prequisites are present in $s$, i.e.\ ${\foo}_{(p_i,e_i) \in c}\, p_i \sqsubseteq s$. 
The compound action $c$ is maximal if no action $a$ can be added such that $c \uplus \multiset{1a}$ is still applicable.
The result $s'$ of applying a compound action $c$ to a state $s$ is the union of the effects of the individual actions: all prerequisites are removed from $s$, and all add lists are inserted, i.e., $s' = (s \uminus p') \uplus e'$ with $p'=\foo_{(p_i,e_i) \in c} \, p_i$ and $e' = \foo_{(p_i,e_i) \in c}\, e_i$.

\myparagraph{Sampling:}
\emph{Probabilistic} MRSs assign a \emph{weight} to each action. 
Given a state $s$, these weights, and the number of possibilities the compound actions can be instantiated, define a distribution of compound actions $p(C \mymid S)$. 
More details are provided in \cite{barbuti_maximally_2011}.
Using this distribution, it is easy to draw sample trajectories: 
Given a state $s$, we calculate all applicable compound actions and their probabilities, sample a single compound action from $p(C \mymid S{=}s)$, and apply it to $s$. 
This process is iterated for the resulting state $s'$. 

\section{Factorizing Distributions over Multisets of Structured Entities}
\label{sec:factorisation}
In this and the following section, we present the main technical contribution: The factorization of distributions of multisets over structured entities, and a probabilistic maximally parallel MRS (PMPMRS) that operates directly on this factorized, parametric representation.

\subsection{Problem Statement}
\label{subsec:problem}
We start by outlining the problem that makes the factorized representation necessary.
For RBSE, we are not interested in \emph{sampling} trajectories of states (as outline above), but in maintaining a distribution of multiset states.

The question is how to efficiently represent such a distribution $p(S{=}s)$. 
If the set of species $\mathcal{E}$ is small and finite, the number of multisets over $\mathcal{E}$ with nonzero support will typically also be small. 
Therefore, we can simply maintain a set of tuples $(s_i,p_i)$ that represent a categorical distribution.
However, if $\mathcal{E}$ is large or even continuous and thus also the number of multisets over $\mathcal{E}$, storing the resulting large or infinite number of tuples $(s_i,p_i)$ becomes infeasible.

The conventional solution for distributions over \emph{metrical} random variables is to use \emph{parametric} distributions that can be represented and manipulated efficiently on the syntactical level, i.e.\ by storing and manipulating just the parameters of the distribution.
For example, a Kalman filter uses ${p(S) \propto \mathcal{N}(\mu,\sigma^2)}$ to represent a distribution over continuous states by storing and manipulating $\mu$ and $\sigma^2$.
Unfortunately, multisets do not necessarily possess a metrical structure that allows to use parametric distributions.
In our approach, we decompose the multisets into a metrical part that allows for parametric distributions, and a remaining, discrete part that that can be represented by a categorical distribution with small (or at least finite) support.
We achieve this by introducing \emph{structured} entities that allow for such a decomposition.

\subsection{Structured Species}
\label{subsec:structured-entities}
An \emph{entity} (a species with internal structure) is a map of property names $\mathcal{K}$ to values $\mathcal{V}$, i.e.\ a partial function $\mathcal{E} = \mathcal{K}\pfun\mathcal{V}$.
This is a necessity for the scenarios we are considering, as they contain entities with multiple, possibly continuous, properties. 
For example, an entity might have a continuous location, e.g.\ a real number.
Using \emph{flat} (unstructured) species, this would require to introduce an infinite number of possible species, and potentially also to an infinite number of actions (as each action prerequisite must be a specific species). 
Using structured entities, the action prerequisites can be expressed more succinctly as constraints on the entities' properties, as described below.
Another reason for using structured entities is that they allow for factorizing the distributions of multisets.

A multiset over entities $\mathcal{E}$ is a state of our MRS. 
Specificaly, we call a multiset $s \in \mathcal{S}$ a \emph{ground state}.
For example, the following state describes a situation in a person tracking scenario where two entities, named Alice and Bob, are at the continuous locations $1.3$ and $2.1$\footnote{For simplicity, the location is a continuous univariate number; in realistic scenarios we might e.g. use 2D locations.}:
\begin{equation}
\multiset{1 \entity{\slot{N}{Alice},\slot{L}{1.3}}, 1 \entity{\slot{N}{Bob},\slot{L}{2.1}}}
\label{eq:ground-state}
\end{equation}

\subsection{Factorising Multiset Distributions}
\label{subsec:factorisation}
Suppose we can decompose the state $s$ into two parts $t$ and $v$, such that there is a bijection from $s$ to tuples $(t,v)$.
%the \emph{context} $c$, describing the actual values of the properties.
Then, a state distribution can be factorized as 
\begin{equation}
p(S)=p(T,V)=p(T)\, p(V \mymid T).
\label{eq:factorisation}
\end{equation}
This idea is independent of multisets and is called Rao-Blackwellization \cite{doucet_rao-blackwellised_2000}.
The question is how to decompose multiset states efficiently, such that the decomposition leads to a more efficient representation of the distribution: 
The factor $p(V \mymid T)$ is handled parametrically, while $p(T)$ has a smaller support than $p(S)$. 

\myparagraph{Decomposition:}
The structured species described in Section~\ref{subsec:structured-entities} allow for such a decomposition: 
We separate the structure of the multiset (how many entities are there, and what are their properties) from the values of the properties.
More formally, the decomposition is performed as follows:
A structure $t \in \mathcal{T}$ (a multiset over $\mathcal{E}_\tau$) is a multiset of entities where the property values are variables, instead of specific values (called \emph{entity structures} $\mathcal{E}_\tau$). The values $v \in \mathcal{W}$ are a list of specific values (of the properties).
For example, the state from Equation~\ref{eq:ground-state} is decomposed into:
\begin{equation}
\begin{split}
t =& \multiset{1\entity{\slot{N}{n_1},\slot{L}{l_1}},1\entity{\slot{N}{n_2},\slot{L}{l_2}}}\\
v =& (\text{Alice},1.3,\text{Bob},2.1)
\end{split}
\label{eq:decomposition}
\end{equation}
Given $t$ and $v$, the state $s$ can be constructed by assuming a canonical order of entities in $t$ (e.g.\ the lexicographic order) and by replacing each variable by the corresponding value in $v$ (i.e\ the i-th variable is replaced by the i-th value).
This describes a bijection between $(t,v)$ and $s$, which allows us to represent $p(s)$ in a factorized way, according to Equation~\ref{eq:factorisation}.

\myparagraph{Distributions of structure and values:}
Performing RBSE inference requires that the distributions $p(T)$ and $p(V  \mymid T)$ can be represented finitely.
For $p(T)$, this is straightforward: 
Given that $p(S)$ is a categorical distribution with finite support, we can also represent $p(T)$ as a finite categorical distribution, simply because multiple elements of $\mathcal{S}$ are mapped to a single element of $\mathcal{T}$.% (also, $p(T)$ will have a smaller support than $p(S)$). 

We assume that $p(V \mymid T)$ is a product of $m$ parametric distributions with parameters $\theta$ such that 
$p(V \mymid T)=\myprod_i p_i(V_i \mymid T,\theta_i)$. 
%The idea here is that each
Every $p_i$ describes the distribution of one or multiple properties and can be represented by the parameters $\theta_i$, and an indicator signifying which parametric form the distribution has. 
We call $\rho(p_i) \in \mathcal{R}$ the \emph{representation} of $p_i$, and $\mathcal{R}$ the representation space. 
For example, $\mathcal{U}(A,B)$ represents an urn without replacement, containing the elements A and B. How to represent the complete value distribution, i.e.\ $\rho(p(V \mymid T))$, will be discussed next.

\myparagraph{Property-Distribution Association:}
The remaining question is how to associate property values in $t$ with the random variables in $p(V \mymid T{=}t)$. At first, this may seem obvious: We order the entities in $t$, and associate the $i$-th property with random variable $v_i$, as suggested by Equation \ref{eq:decomposition}. 
However, this is not sufficient in many cases, as there might be \emph{non-local} dependencies of multiple values: For example, in Equation~\ref{eq:decomposition}, the distributions of the names of both entities are not independent, assuming that all names are unique. Thus, we must make sure that exactly the name properties are distributed according to a joint distribution (e.g.\ an urn without replacement). 
To succinctly represent which property values are associated with which distributions, we propose a labeling mechanism that provides this association. We introduce this mechanism by describing $\mathcal{E}_\tau$ and the representation of the value distribution, $\rho(p(V \mymid T))$ in more detail.

An \emph{entity structure} is a map from property names $\mathcal{K}$ to labels $\mathcal{D}$, i.e.\ $\mathcal{E}_\tau = \mathcal{K} \pfun \mathcal{D}$. The distribution $p(V \mymid T)$ is represented as a map of labels $\mathcal{D}$ to distribution representations $\mathcal{R}$. Note that these labels are essentially pointers.
We call the representation of $p(V \mymid T)$ the \emph{context} $c_t = \rho(p(V \mymid T))$ of $t$.

This mechanism allows us to easily represent correlations of properties, even when the properties belong to different entities, without the need to define an order of the entities in $t$. This is illustrated in the following example.

\myparagraph{Example:}
Suppose we know that two persons are present in a situation, and we have normally distributed location estimates for both persons (e.g.\ based on a measurement), but we do not know which specific person corresponds to which location estimate. This situation can be described by the following factorized state representation, i.e.\ the pair of structure $t$ and context $c_t$:
\begin{align}
\begin{split}
 t  &= \multiset{1 \entity{\slot{N}{\mathbf{N}},\slot{L}{\mathbf{L_1}}}, 1 \entity{\slot{N}{\mathbf{N}},\slot{L}{\mathbf{L_2}}}} \\
% c_t &= \context{\density{\mathbf{N}}{\mathcal{U}(\text{A,B})},
% \density{\mathbf{L_1}}{\mathcal{N}(1,2.1)},
% \density{\mathbf{L_2}}{\mathcal{N}(2,1,1.5)}}\\
 c_t &= \context{
 \density{\textbf{N}}{\mathcal{U}(\text{A,B})},
 \density{\mathbf{L_1}}{\mathcal{N}(1.3, 2.0)},
  \density{\mathbf{L_2}}{\mathcal{N}(2.1, 1.0)}}\\
\end{split}
\label{eq:state-ab}
\end{align}
Note how in the example, we see how the same distribution $\textbf{N}$ can be referenced multiple times in $t$, which means that the corresponding properties are distributed according to a joint distribution. On the other hand, properties that reference \emph{different} distributions are independent. In the example, the name of an entity is independent of the entity's location.

\myparagraph{Exchangeability:}
Using the labeling mechanism, we do not rely on an order of entities, i.e.\ the order of the entities in $t$ is arbitrary. Therefore, we require all joint distributions in the context to be \emph{exchangeable}, i.e. $p(x_1,x_2)=p(x_2,x_1)$. 
%For example, uniform (non-weighted) urns with and without replacement are exchangeable.
From the view of Lifted Inference, the context thus represents an exchangeable decomposition \cite{niepert2014tractability} of the value distribution. This property is the reason that allows efficient filtering, as will be explained in Section~\ref{sec:rbse}.

\myparagraph{Lifted State:}
Using our techniques, we can represent the categorical distribution $p(S)=p(T,V)$ by: 
(i) A categorical distribution of structures $p(T)$ (i.e. a set of tuples $(t,p)$) and 
(ii) for each $t$, a context $c_t$, representing $p(V \mymid T)$.
Instead of storing a set of tuples $(t,p)$ and a context $c_t$ for each $t$, equivalently, we can directly store a set of triples $(t,p,c_t)$. However, this is simply a categorical distribution of pairs $(t,c_t)$. Thus, the distribution $p(S)$ can be represented by a categorical distribution of such pairs $(t,c_t)$. We call $l=(t,c_t)$ a \emph{lifted state}, and $p(L)$ a \emph{lifted state distribution}.

Each lifted state $l$ describes a distribution over $\mathcal{S}$ where all $s$ have the same structure $t$ and all $v$ are distributed according to $p(V \mymid T{=}t)$. 
Note that the structures $t$ in a lifted state distribution $p(L)$ need not be distinct (due to splitting, see Section~\ref{subsec:splitting}). Thus, a lifted state distribution $p(L{=}l)$ with $l=(t,c_t)$, $\rho(p_i(V_i \mymid t)) \in \text{range}\, c_t $ and $p(V \mymid t)=\Pi_i p_i(V_i \mymid t,c_t)$ describes a distribution of states $s$ as follows\footnote{For $p(V \mymid T)=\myprod_i p_i(V_i \mymid T)$, we assume that the $V_i$ are assigned to the $p_i$ according to their labels: All $V_i$ that have label $d_j$ in $t$ are distributed according to the distribution with label $d_j$ in $c_t$.}:
\begin{align}
\begin{split}
&p(S{=}s)=p(T{=}t,V{=}v) =\\
&\sum_{\{l_i=(t_i,c_i) | t_i=t\}} p(L{=}l_i) \, p(V{=}v \mymid t_i)
\end{split}
\label{eq:l-to-s}
\end{align}

\section{Lifted Filtering via Exchangeable Decomposition}
\label{sec:rbse}

In the following, we present a RBSE algorithm that uses the factorized multiset distribution to represent the current state distribution. 
Given a prior distribution of states $p(S_t \mymid y_{1:t})$, the calculation of the posterior distribution after observing $y_{t+1}$, i.e.\ $p(S_{t+1} \mymid y_{1:t+1})$ can be decomposed into the following two steps:
The \emph{predict} step calculates the distribution after applying the \emph{transition model} $p(S_{t+1} \mymid S_t)$, i.e. $p(S_{t+1} \mymid y_{1:t}) = \sum_{s_t} p(S_{t+1} \mymid S_t{=}s_t)\, p(S_t{=}s_t \mymid y_{1:t})$.
Afterwards, the posterior distribution is calculated by employing the \emph{observation model} $p(y_{t+1} \mymid S_{t+1})$:
\begin{equation}
p(S_{t+1} \mymid y_{1:t+1}) = \frac{p(y_{t+1} \mymid S_{t+1})\, p(S_{t+1} \mymid y_{1:t})}{p(y_{t+1}\mymid y_{1:t})}
\end{equation}

Interestingly, these steps can be performed directly on the lifted states. This is possible because for multiset rewriting, it is only necessary to know \emph{how many} entities have a certain property, but their specific order or identity is not relevant.
It has been shown that exactly such exchangeability properties (reflected by exchangeability of the value distribution) allow efficient Lifted Inference \cite{niepert2014tractability}.

\subsection{Predict}
\label{subsec:state-dynamics}
For the predict step, we need to define the transition model $p(S_{t+1} \mymid S_t)$. 
The dynamics of the system is described in terms of compound actions, as introduced in Section~\ref{sec:background}.
However, we have to account for the structured entities (leading to more complex preconditions and effects) and the lifted state representation (requiring \emph{splitting} when the preconditions are indeterminate, explained in Section~\ref{subsec:splitting}).
In the following, these concepts are introduced more formally.

\myparagraph{Actions} describe the behavior of the entities. An action is a pair $(c,e,w)$ consisting of a precondition list $c \in \mathcal{C}$ and an effect function $e \in \mathcal{F}$.
A precondition list is a list of constraints on entity structures and their corresponding values, i.e. boolean functions: $\mathcal{C} = [\mathcal{E}_\tau \times V \rightarrow \{\top,\bot\} ]$.
The idea of applying an action to a lifted state is to \emph{bind} entities to the preconditions.
Specifically, one entity is bound to each element in the precondition list, and entities can only be bound when they satisfy the corresponding constraint.  
 The effect function then manipulates the state based on the bound entities (by inserting, removing, or manipulating entities or the distributions stored in the context).
We call such a binding \textbf{action instance}, i.e.\ an action instance is a pair of an action and a list of entity structures.
Entity structures can be indeterminate regarding a precondition, as some values $v$ drawn from $p(V \mymid T)$ (represented by $c_t$) may satisfy the precondition, while others do not. This case requires \emph{splitting}, described in Section~\ref{subsec:splitting}.
 For now, we assume that the preconditions are determinate.

\myparagraph{A Compound Action} $k \in \mathcal{K}$ is a multiset of action instances. 
It is applied to a state by composing the effects of the individual action instances.
In the following, we are only concerned with applicable maximal compound action (AMCAs, see Section~\ref{sec:background}), which define the transition model.

\myparagraph{Compound action probabilities:}
Our system is probabilistic, which means that each AMCA is assigned a probability.
In general, any function from the AMCAs to positive real numbers which integrates to one is a valid definition of these probabilities, that might be plausible for different domains. 
Here, we use the probabilities that arise when each entity independently chooses which action to participate in (which is the intended semantics for the activity recognition problems we are concerned with).
In this case, the probability of each compound action is based on the number of options specific entities can be bound to the preconditions of the actions, and a weight for each action. 
For details, we refer to \cite{barbuti_maximally_2011}.

\myparagraph{Transition model:}
The distribution of the AMCAs define the distribution of successor states, i.e.\ the \emph{transition model}. 
The successor states of $l$ are obtained by applying all AMCAs to $l$.
The probability of each successor state $l'$ is the sum of the probabilities of all AMCAs leading to $l'$:
\begin{equation}
p(L'{=}l' \mymid L{=}l) = \sum_{\{k | apply(k,l)=l'\}} p(K{=}k \mymid L{=}l)
\label{eq:trans-model}
\end{equation} 

Thus, given a prior lifted state distribution and a set of actions, the following steps have to be performed to obtain the posterior lifted state distribution: 
(i) Split the lifted states until all preconditions are determinate, see Section~\ref{subsec:splitting}, 
(ii) compute all action instances of each action, 
(iii) compute all AMCAs and their probabilities, 
(iv) apply all AMCAs to the lifted state, 
(v) calculate the probabilities of the resulting successor states.

Step (ii) can be solved by simple forward constraint satisfaction. 
Step (iii) can also be implemented by a search-based approach, i.e.\ building the compound actions by incrementally adding all applicable action instances. Additional efficiency can be gained by noting that multiset insertion is commutative. Therefore, we can define an arbitrary order of the action instances and when incrementally building the compound actions only insert instances that are ``not smaller'' than the last inserted action instance.

\subsection{Update}
Given a state distribution $p(S \mymid y_t)=p(T,V  \mymid  y_{t})$ (which might be represented by a lifted state representation via Equation \ref{eq:l-to-s}), the update step can be decomposed into the update of the structure, and the update of the value distribution. 
This becomes obvious by applying the chain rule and Bayes' theorem to the distribution:
\begin{align}
\begin{split}
 p(S \mymid y_{t}) = & p(T,V  \mymid  y_{t}) \\
 = & p(T \mymid y_{t}) \, p(V \mymid y_{t},T)\\
 = & \underbrace{\frac{p(y_{t} \mymid T) p(T)}{p(y_{t})}}_\text{(a)}  \underbrace{\frac{p(y_{t} \mymid V,T)p(V \mymid T)}{p(y_{t} \mymid T)}}_\text{(b)} 
 \end{split}
\end{align}

The factor (a) corresponds to the update of the structure, and (b) is the update of the value distribution.
The values $p(y_t)$ and $p(y_t \mymid T)$ are normalization factors that can be obtained by marginalization.
Both factors of the update can be computed on the lifted state representation:
Step (a) means that the weight of each structure (i.e. the weight of the corresponding lifted state) is multiplied by observation probability $p(y_{t+1} \mymid T)$.
Step (b) corresponds to syntactically manipulating of each context $c_t$, i.e. modifying the parametric distribution $p(V \mymid t)$. 
For example, if $P(V \mymid T)$ and $p(y_{t+1} \mymid V,T)$ follow normal distributions, this corresponds to the standard Kalman filter update.

When $p(V \mymid y_{t+1},T)$ can no longer be represented by a product of exchangeable parametric distributions, we have to \emph{split} the lifted state, described in Section \ref{subsec:splitting}.
For example, regarding the situation in Equation~\ref{eq:state-ab}, when observing a speficic person (say, Alice) at a specific location, the name and location are no longer independent.

\subsection{Splitting}
\label{subsec:splitting}
During action instance computation and during the update step, two closely related problems arise:
(1) While computing action instances, a precondition can be satisfied for some ground states $s$ represented by a lifted state $l$, and not satisfied by others which are also represented by $l$. 
(2) Due to observations, the resulting ground state distribution previously represented by a lifted state may no longer be representable by a single lifted state.
In general, the ground states form partitions based on whether a constraint is satisfied or not (in action instance computation) or whether they can be represented exactly by a single lifted state using an exchangeable parametric distribution for $p(V \mymid T)$ in the update step.

We want to compute lifted states that describe the partitions, without requiring a complete enumeration of all ground states first.
This is done by manipulating the lifted states by an operation called \emph{splitting}.
More specifically, splitting a lifted state $l$ results in a set of lifted states $L$ such that: (1) $L$ describes the same distribution of ground states as $l$, and (2) for each $l_i \in L$, all ground states $s_i$ described by $l_i$ lie in the same partition.
How splitting is done exactly depends on the parametric form of $p(V)$. In the following, we describe splitting for urns without replacement on equality constraints:

Suppose we want to split the property $q$ on the constraint ``$q=v$''. Intuitively, the strategy is to ground the values of the specific property $q$. Suppose the values of $q$ are distributed according to an urn without replacement with $n$ different values, $\mathcal{U}(v_1,\dots,v_n)$. We split this situation into $n$ worlds, where in world $i$, $q$ has value $v_i$. For example, splitting the state in Equation~\ref{eq:state-ab} results in two lifted states, one where Alice is at the location distributed according to $L_1$, and one where Alice is at the location distributed according to $L_2$.
Splitting into $n$ lifted states (instead of two lifted states, one where the precondition is satisfied for $e$ and one where it is not satisfied) is necessary to preserve the independence and exchangeability properties of the remaining urn. The general procedure for splitting urns is shown in Algorithm~\ref{alg:split-urns}.
Note that splitting only affects the specific property that we split on, but other properties are still represented in lifted form.
The approach is conceptually similar to splitting parametric factors in First-order Variable Elimination \cite{poole_first-order_2003}.

Splitting rules for other kinds of parametric distributions can be developed by following the same idea. In general, we can split a distribution when a corresponding conditional distribution (conditioned on the constraint that we split on) is still exchangeable and can still be represented parametrically.

\newcommand*\Let[2]{\State #1 $\gets$ #2}
\begin{algorithm}[tb]
  \caption{For lifted state $l$, split the property $q \in \mathcal{K}$ with label $d \in \mathcal{D}$, distributed according to urn $u$.}
    \label{alg:split-urns}
  \begin{algorithmic}[1]
      \Function{split-urn}{$l{=}(t,c)$,$q$,$d$,$u$}
    \For{value $v$ in $u$}
        \Let{$u'$}{$u$ without $v$}
        \Let{$e'$}{$e \oplus \entity{\slot{q}{d'}}$}
        \Let{$c'$}{$c \oplus \context{\density{d}{u'}, \density{d'}{\delta_v}}$}
   	    \Let{$t'$}{$t \uminus \multiset{1e} \uplus \multiset{1e'}$}
		\Let{$l'$}{$(t',c')$}
    		\Let{$p(l')$}{Probability of $v$ in $u$}
  	\EndFor
    	\State \Return{Categorical Distribution $p(L')$}
    \EndFunction
  \end{algorithmic}
\end{algorithm}

% !TEX root = ../P20171110-IJCAI_LiMa-SL.tex

\section{Qualitative Examples}
\label{sec:evaluation}
In this section, we provide an intuition for the qualitative effect of the lifted state representation, by comparing LiMa with an \emph{exact propositional} filtering approach that maintains $p(S)$ by complete enumeration. We compare the the cardinality $p(L)$ (handled by LiMa) and $p(S)$ -- which is an indicator of space and time complexity. 
 We do not compare LiMa with other approaches, as they are either approximate, or cannot handle identifying observations at all.

\begin{figure}[tb]
\centering
\includegraphics[scale=0.4]{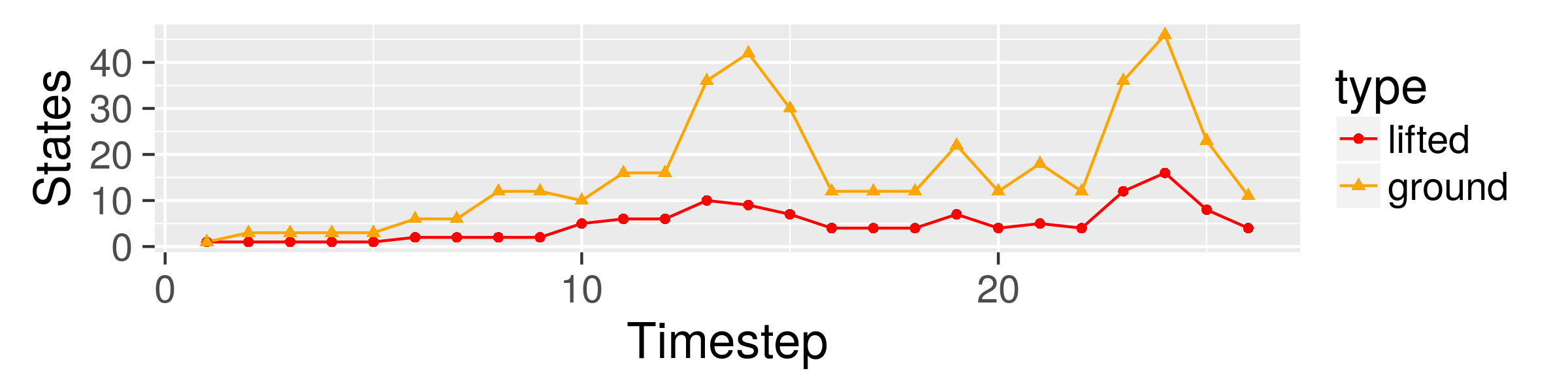}
\caption{Number of lifted and grounded states during inference for the first scenario.}
\label{fig:abc}
\end{figure}

\begin{figure}[tb]
\centering
\includegraphics[scale=0.4]{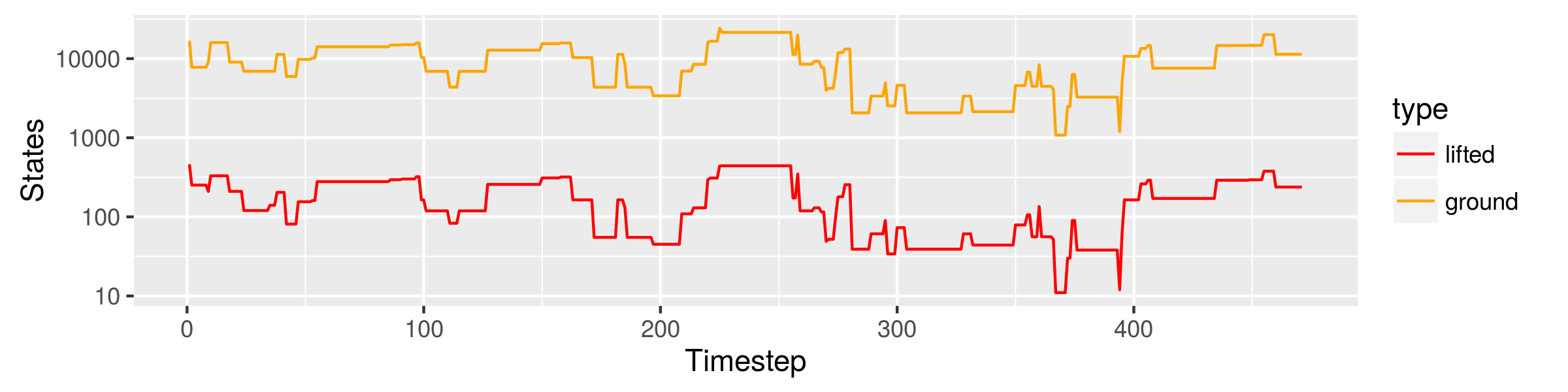}
\caption{Number of lifted and grounded states during inference for the second scenario.}
\label{fig:office-over-time}
\end{figure}

\begin{figure}[tb]
\centering
\includegraphics[scale=0.6]{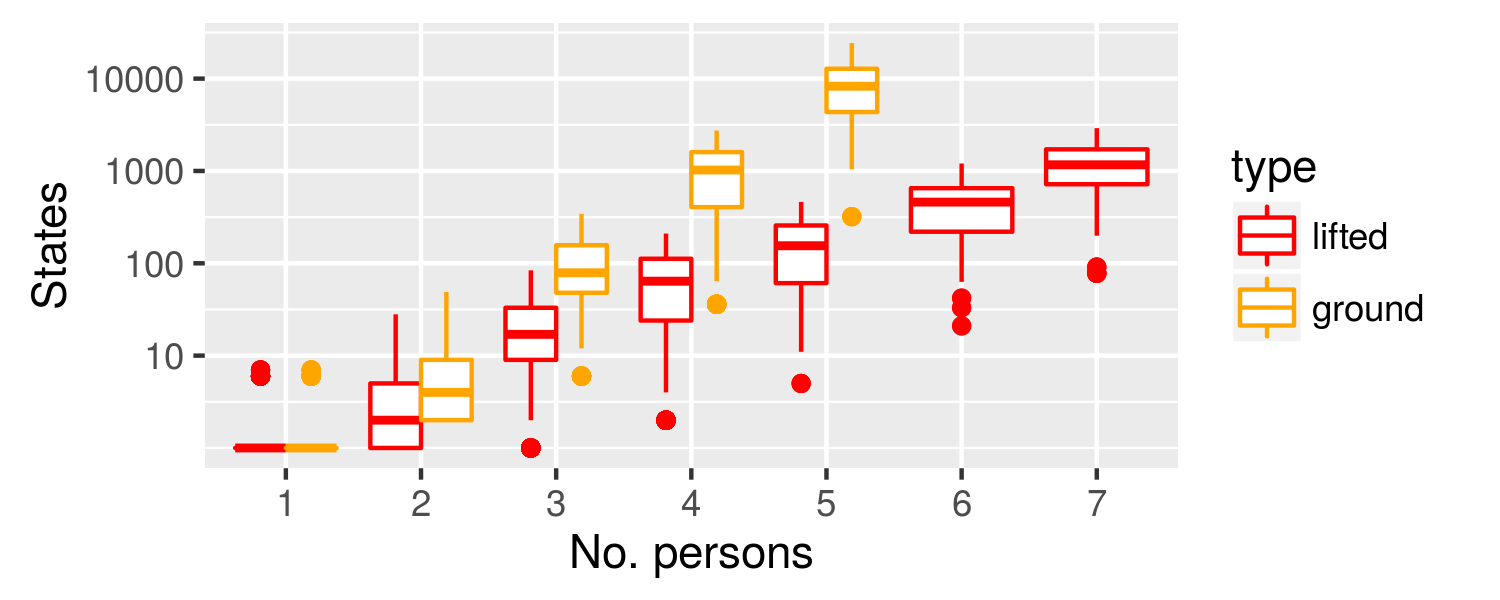}
\caption{Mean number of states for second scenario for different number of persons.}
\label{fig:office}
\end{figure}

We use two scenarios from the HAR domain for evaluation.
The first scenario \cite{schroder_sequential_2017} consists of simulated sensor data of three persons acting in an office environment observed by presence sensors. For each person, their name, location and whether they hold an object is modeled.
The simulated sensor data do not reveal the \emph{identity} of the persons nor the number of persons per location.
At timestep $t=10$, an \emph{identifying} observation is been made (Alice is at the printer), resulting in a split of the lifted states on the predicate ``Name=Alice''.

The second scenario is a person tracking task with anonymous sensors (similar to the previous scenario), but uses real sensor data of PIR and light switch sensors.
The data consists of 35 observation sequences of 1 to 7 agents (5 each) who move between 14 rooms in an office.
The dataset is available at \cite{kasparick2013probabilistic}.
We model both scenarios by entities whose name properties are distributed according to a finite urn.
Both scenarios have a compound action semantics, as all persons can simultaneously move between observations.

Figures~\ref{fig:abc} and \ref{fig:office-over-time} show the number of states necessary to represent the posterior $p(S_t)$ over time. For both scenarios, using the lifted state representation, the number of states represented explicitly is several times smaller. 
Note that the ground and the lifted states represent exactly the same distribution (see Equation~\ref{eq:l-to-s}).
Figure~\ref{fig:office} shows the mean number of states occurring during inference for the second scenario, for different numbers of agents. 
Here, we see that the effect gets more pronounced for larger number of agents. 
The reason for this is the increase of the number of ground states due to the factorial number of entity-name associations (that are all represented by a single lifted state).

The number of explicit state representations linearly corresponds to runtime, as compound action computation (which is the most computationally expensive part of the filtering algorithm) has to be performed individually for each state.
Thus, the empirical results clearly demonstrate that LiMa can be order of magnitude faster.

%%%%%%%%%%%%%%%%%%%%%%%%%%%%%%%%%%%%%%%%%%%%%%%%%%%%%%%%%%%%%%%%%%%%%%%%%%%%
%%%%%%%%%%%%%%%%%%%%%%%%%%%%%%%%%%%%%%%%%%%%%%%%%%%%%%%%%%%%%%%%%%%%%%%%%%%%

\section{Related Work}
Another class of approaches concerned with performing inference over compact representations of probability distributions is known as \emph{Lifted Probabilistic Inference} \cite{de_raedt_statistical_2016}. 
Lifted Inference can be seen as decomposing a distribution into exchangeable components and handling sufficient statistics of them \cite{niepert2014tractability}. The exchangeable decomposition is made explicit in our approach -- each factor of the value distribution is exchangeable. We are then also only interested in sufficient statistics, namely \emph{how many} entities have a certain property.

Ideas of Lifted Inference have been applied to RBSE (i.e.\ to dynamic domains) in the Relational Kalman Filter \cite{choi_learning_2015}, an approach that is restricted to a Gaussian state distribution and a linear transition model.

The primary reason that makes it difficult to directly apply Lifted Inference algorithms to MRSs are the hard constraints that are present in the transition semantics: which entity can perform which actions, the fact that each entity must perform exactly one action etc.
Interestingly, computing compound actions is a special case of Lifted Weighted Model Counting \cite{gogate2012probabilistic}, that additionally considers such hard constrains.

There are a number of other approaches that aim at finding compact descriptions of sets of states in dynamic systems, like \emph{Relational POMDPs} \cite{sanner_practical_2009-1} and \emph{Logical Filtering} \cite{shirazi2011first}. They employ situation calculus to describe states and actions, but are not explicitly concerned with efficient filtering.
The idea of using independent factors for RBSE in multi-agent settings is also explored by 
\cite{pfeffer_factored_2009}, 
 but this approach needs to multiply the factors and then sample from the joint, whereas LiMa only resorts to the joint distribution when necessary.

An RBSE algorithm that, similar to LiMa, uses state descriptions that each represents a set of specific states is the \emph{Relational Particle Filter} \cite{nitti_particle_2013}. 
It uses states where some variables are described by specific values and others are represented by parametric distributions. 
Similar to LiMa, the approach can handle continuous and infinite domains. In cases where we require a split, the algorithm samples from the corresponding distributions, instead of manipulating the exact distributions on a parametric level.

\emph{Stochastic Relational Processes} (SRPs) \cite{thon_stochastic_2011} use causal probabilistic logic to describe the transition model of a RBSE algorithm, i.e.\ by a set of probabilistic precondition-effect rules. Opposed to the factorized multiset states used in LiMa, SRPs use a ground state representation -- although a part of the transition model can be calculated in a lifted way (for calculating the successor states, not all ground rules need to be generated).

% !TEX root = ../P20171110-IJCAI_LiMa-SL.tex

\section{Conclusion}
We presented LiMa, a recursive Bayesian state estimation (RBSE) algorithm that uses a Probabilistic Maximally Parallel Multiset Rewriting System to model the underlying state dynamics.
It uses a \emph{factorized} representation for distributions of multisets and exploits exchangeability to perform the complete RBSE cycle on this compact representation, without needing to sample from or to enumerate all ground states.
Empirical evidence in two activity recognition domains shows that LiMa needs to represent a much lower number of states explicitly.

There are scenarios (e.g.\ containing many identifying observations) that might require repeated splitting until the state representation is completely grounded, which is a well-known problem in the filtering literature \cite{boyen1998tractable}.
Therefore, future research will concentrate on methods that allow to maintain the factorized representation when the independence assumptions hold only approximately, methods that re-introduce factorized representations (similar to \emph{merging} in Lifted Inference, e.g.\ \cite{kersting_counting_2009}), and using other representations for $p(V{|}T)$, like Sum-Product Networks \cite{poon_sum-product_2011} or Exchangeable Variable Models \cite{niepert_exchangeable_2014}.
A further research goal is to investigate whether the multiplicities of entities can also be represented by parametric distributions, which would lead to an even more compact representation of distributions.

\section*{Acknowledgements}
This work was supported by the \emph{Deutsche Forschungsgemeinschaft} (DFG), grant number INST 264/128-1 FUGG.

%%%%%%%%%%%%%%%%%%%%%%%%%%%%%%%%%%%%%%%%%%%%%%%%%%%%%%%%%%%%%%%%%%%%%%%%%%%%%%%
% BIBLIOGRAPHY                                                                %
%%%%%%%%%%%%%%%%%%%%%%%%%%%%%%%%%%%%%%%%%%%%%%%%%%%%%%%%%%%%%%%%%%%%%%%%%%%%%%%

\appendix

%% The file named.bst is a bibliography style file for BibTeX 0.99c
\bibliographystyle{Template/named}
\bibliography{P20171110-IJCAI_LiMa-SL}

\end{document}